\documentclass{article}
\usepackage{spconf,amsmath,graphicx}
\usepackage{amsfonts,amssymb} 
\usepackage{svg}
\usepackage{multirow}
\usepackage{subfigure}
\usepackage{algorithm}
\usepackage{algorithmic}
\usepackage[normalem]{ulem}
\usepackage{colortbl}
\usepackage{bbding}
\usepackage{setspace}
\usepackage{pifont}
\usepackage{wasysym}
\usepackage{amssymb}
\usepackage{caption} 
\usepackage{lipsum}

\newcommand{\cmark}{\ding{51}}%
\newcommand{\xmark}{\ding{55}}%

\title{Consistent training and decoding for End-to-End Speech Recognition using Lattice-Free MMI}
\name{Jinchuan Tian$^{1}$\thanks{This work is done when Jinchuan Tian is an intern in Tencent AI lab.},
      Jianwei Yu$^{2}$\thanks{Paper submitted to ICASSP 2022.}, 
      Chao Weng$^2$,
      Shi-Xiong Zhang$^2$,
      Dan Su$^2$,
      Dong Yu$^2$, 
      Yuexian Zou$^{1,*}$\thanks{* Corresponding author.}} 
\address{$^1$ADSPLAB, School of ECE, Peking University, Shenzhen, China \\
         $^2$Tencent AI Lab}
\begin{document}

\ninept
\maketitle
\begin{abstract}
\noindent Recently, End-to-End (E2E) frameworks have achieved remarkable results on various Automatic Speech Recognition (ASR) tasks. However, Lattice-Free Maximum Mutual Information (LF-MMI), as one of the discriminative training criteria that show superior performance in hybrid ASR systems, is rarely adopted in E2E ASR frameworks. 
In this work, we propose a novel approach to integrate LF-MMI criterion into E2E ASR frameworks in both training and decoding stages. The proposed approach shows its effectiveness on two of the most widely used E2E frameworks including Attention-Based Encoder-Decoders (AEDs) and Neural Transducers (NTs). 
Experiments suggest that the introduction of the LF-MMI criterion consistently leads to significant performance improvements on various datasets and different E2E ASR frameworks. 
The best of our models achieves competitive CER of 4.1\% / 4.4\% on Aishell-1 dev/test set; we also achieve significant error reduction on Aishell-2 and Librispeech datasets over strong baselines. Code available\footnote{https://github.com/jctian98/e2e\_lfmmi}. 
\end{abstract}
\section{Introduction}
\label{sec:intro}



In the past few years, the performance of Automatic Speech Recognition (ASR) systems is greatly advanced due to the prosperity of End-to-End (E2E) frameworks\cite{e2easr_survey}. Currently, Attention-Based Encoder-Decoders (AEDs)\cite{las, lasctc} and Neural Transducers (NTs)\cite{rnnt} are two branches of the most popular frameworks in E2E ASR. 
In general practice, training criteria like Cross-Entropy (CE), Connectionist Temporal Classification (CTC)\cite{ctc} and Transducer Loss\cite{rnnt} are adopted in AEDs and NTs. However, all of the three criteria try to directly maximize the posterior of the transcription given acoustic features but ignore other competitive hypotheses. 

Recently, motivated by the success of discriminative training criteria (e.g., MPE\cite{mpe,povey_phd}, sMBR\cite{mpe, povey_phd, smbr} and MMI\cite{mpe, povey_phd,lfmmi_16, lfmmi_18}) in hybrid ASR systems, there are several attempts to incorporate them into E2E frameworks. 
In \cite{mwer_las, mbr_las, mbr_rnnt, mwer_rnnt}, the word-level Minimum Bayesian Risk (MBR) criterion is applied to AEDs\cite{mwer_las, mbr_las} and NTs\cite{mbr_rnnt, mwer_rnnt} during system training and achieves competitive recognition performance. 
In addition, MMI and MBR criteria \cite{sa_mmi,sa_mbr} that are dedicated in discriminating hypotheses from different speakers are also adopted in speaker-attributed ASR systems. 
However, there are still some deficiencies in current approaches. First, MBR-based methods\cite{mwer_las, mbr_las, mbr_rnnt, mwer_rnnt} work in a two-stage style: they require a trained model for initialization and on-the-fly decoding to generate hypotheses for discrimination, which results in complex working pipeline, low training efficiency and exceeded memory consumption. Also, current methods for E2E ASR systems only use discriminative training criterion in the training process, which results in a mismatch between training and decoding. 

In this work, we propose to integrate LF-MMI into E2E ASR systems, specifically AEDs and NTs. Unlike the methods aforementioned, the proposed method works in a one-stage style and adopts the LF-MMI criterion consistently in both system training and decoding. 
In the proposed method, the E2E ASR systems are optimized by both LF-MMI and other non-discriminative objective functions in training. During decoding, evidence provided by LF-MMI is consistently used in either beam search or rescoring. In terms of beam search, MMI Prefix Score is proposed to evaluate partial hypotheses of AEDs while MMI Alignment Score is adopted to assess the hypotheses proposed by NTs. 
In terms of rescoring, the N-best hypothesis list generated without LF-MMI is further rescored according to the LF-MMI scores. To verify the effectiveness of our method, experiments are conducted on both Mandarin (Aishell-1, Aishell-2) and English (Librispeech) datasets. Our experiments suggest that adding LF-MMI as an additional criterion in training can improve the recognition performance. Moreover, decoding with LF-MMI scores will further improve the performance of these systems. Among various attempts, the best of our models achieves CER of 4.1\%  and 4.4\% on Aishell-1 dev/test set. To the best of our knowledge, this is the state-of-the-art result of NT systems on Aishell-1. We also achieve 0.5\% and 0.3\% character / word error rate (CER/WER) reduction absolutely on Aishell-2 \textit{test-ios} set and Librispeech \textit{test-other} set respectively.

To conclude, we propose a novel approach to integrate discriminative LF-MMI criterion into E2E ASR systems not only in system training but also in the decoding process. Specifically, three decoding algorithms are proposed to incorporate LF-MMI scores into both first-pass decoding and second-pass rescoring for AED and NT frameworks. To the best of our knowledge, this paper is among the first works to apply LF-MMI criterion to E2E ASR systems that maintains the consistency between training and decoding. In contrast, previous works \cite{mwer_las, mbr_las, mbr_rnnt, mwer_rnnt} only consider discriminative criteria in training.
\vspace{-7pt}
\section{LF-MMI training}
In ASR, the MMI criterion is used to discriminate the correct hypothesis from all hypotheses by maximizing the ratio as follows:
\begin{equation}
\setlength\abovedisplayskip{0cm}
\setlength\belowdisplayskip{0cm}
\begin{aligned}
    \log P_{\tt{MMI}}(\mathbf{W}|\mathbf{O})
    &=\log\frac{P(\mathbf{O}|\mathbf{W})P(\mathbf{W})}{\sum_{\mathbf{\bar{W}}}P(\mathbf{O}|\mathbf{\bar{W}})P(\mathbf{\bar{W}})}
\end{aligned}
\label{eq_mmi}
\end{equation}
where $\mathbf{O}$, $\mathbf{W}$ and $\bar{\mathbf{W}}$ represent the acoustic feature sequence, transcription and any possible hypothesis respectively. 
However, directly enumerating $\bar{\textbf{W}}$ is almost impossible in practice. 
Thus, Lattice-Free MMI\cite{lfmmi_16, lfmmi_18} is proposed to approximate the numerator and denominator in Eq.\ref{eq_mmi} by \textit{forward-backward} algorithm on two Finite-State Acceptors (FSAs). The log-posterior of $\mathbf{W}$ is then converted into the ratio of likelihood given the graphs and $\mathbf{O}$ as follow:
\begin{equation}
\setlength\abovedisplayskip{0cm}
\setlength\belowdisplayskip{0cm}
\begin{aligned}
    \log P_{\tt{LF-MMI}}(\mathbf{W}|\mathbf{O}) 
    \approx \log\frac{P(\mathbf{O}|\mathbb{G}_{num})}{P(\mathbf{O}|\mathbb{G}_{den})}
\end{aligned}
\label{eq_mmi_graph}
\end{equation}
where $\mathbb{G}_{num}$ and $\mathbb{G}_{den}$ denotes the FSA numerator graph and denominator graph respectively. 
Unlike the lattice-based method, the denominator graph in LF-MMI is built from a phone-level language model and is identical to all utterances, which avoids the pre-decoding process before training and could be used from scratch. 
The mono-phone modeling units are adopted in this work, as a large number of modeling units (e.g. Chinese characters, English BPEs) makes the denominator graph computationally expensive and memory-consuming. 


\subsection{LF-MMI Training in E2E Systems}
\begin{figure}[htpb]
    \centering
    \subfigure[Attention-Based Encoder-Decoder (AED)]{
    \includegraphics[width=7cm]{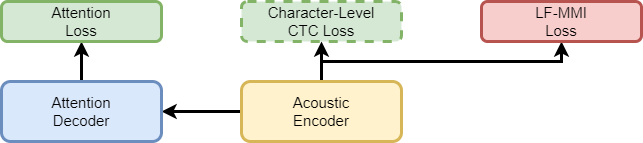}
    }
    \quad
    \centering
    \subfigure[Neural Transducer (NT)]{
    \includegraphics[width=7cm]{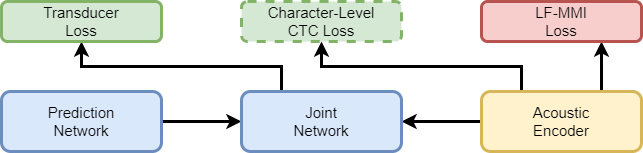}
    }
    \caption{Diagram about how LF-MMI criterion is integrated into AEDs and NTs during training. Character-Level CTC Loss is optional. All losses are optimized as a weighted sum.}
    \label{fig_loss}
\end{figure}

As shown in Fig.\ref{fig_loss}, the LF-MMI criterion is used as an auxiliary criterion to optimize the acoustic encoder in both AED and NT frameworks. The global training objective to minimize is formulated as:
\begin{equation}
    \label{weighted_j}
    \mathbf{J} = 
                 \left\{
                 \begin{aligned}
                 &(1-\beta)\cdot\mathbf{J}_{\tt{ATT}} + \beta\cdot\mathbf{J}_{\tt{CTC}}                                     \\
                 &\mathbf{J}_{\tt{NT}}                                        \\
                 \end{aligned}
                 \right\} - \alpha \cdot \log P_{\tt{MMI}}(\mathbf{W}|\mathbf{O}) 
\end{equation}
where $\mathbf{J}_{\tt{ATT}}$, $\mathbf{J}_{\tt{CTC}}$ and $\mathbf{J}_{\tt{NT}}$ denote the Attention loss, CTC loss and Transducer loss respectively. Empirically, the weight of LF-MMI criterion $\alpha$ is set to 0.3 and 0.5 for AEDs and NTs respectively. As regularization is necessary for LF-MMI\cite{lfmmi_16}, Character-Level CTC is found as an ideal regularization and is optionally adopted with the same weight as LF-MMI criterion during training.

\vspace{-17pt}
\section{LF-MMI Decoding}
\vspace{-5pt}
To tightly integrate the training and decoding process, we also integrate the LF-MMI criterion in the decoding process. 
In this section, MMI Prefix Score ($S^{\tt{pref}}_{\tt{MMI}}$) and MMI Alignment Score ($S^{\tt{ali}}_{\tt{MMI}}$) are proposed to integrate LF-MMI scores into beam search of AEDs and NTs respectively. To apply our method to spelling languages like English, A look-ahead strategy is subsequently provided. In addition, we also propose a rescoring method using LF-MMI. 
\vspace{-10pt}
\subsection{Beam Search for AEDs}
Assume $\mathbf{H}(\mathbf{W}_{1}^u)$ is the set of all possible hypotheses that start with a partial hypothesis $\mathbf{W}_{1}^u=[<sos>, w_1, ..., w_u]$. The goal of decoding for AED systems is to find the most probable hypothesis $\hat{\mathbf{W}}$ in  $\mathbf{H}([<sos>])$ given the acoustic feature sequence $\mathbf{O}=[o_1, ..., o_T]$ and $\mathbf{W}_{1}^u=[<sos>]$.
\begin{equation}
\setlength\abovedisplayskip{0cm}
\setlength\belowdisplayskip{0cm}
\label{concept_search}
    \hat{\mathbf{W}} = \mathop{\arg\max}_{\mathbf{W}\in \mathbf{H}([<sos>])}P(\mathbf{W}|\mathbf{O})
\end{equation}
Normally, this \textit{maximize-a-posterior} process is approximated by the beam search. 
Assume $\mathbf{\Omega}_u$ is the set of active partial hypotheses with length $u$. 
Then $\mathbf{\Omega}_{u}$ is recursively generated by expanding each partial hypothesis in $\mathbf{\Omega}_{u-1}$ and pruning those expanded partial hypotheses with lower scores. 
This iterative process would terminate once the stopping condition is met. 
Typically, we set $\mathbf{\Omega}_0=\{[<sos>]\}$ while all hypotheses in any $\mathbf{\Omega}_u$ that end with $<eos>$ would be moved to a finished hypothesis set $\mathbf{\Omega}_F$ for final decision. The computation of partial scores is the basis of beam search. Partial score $\alpha(\mathbf{W}_{1}^u, \mathbf{O})$ of a partial hypothesis $\mathbf{W}_{1}^u$ is recursively computed as:
\begin{equation}
\setlength\abovedisplayskip{1pt}
\setlength\belowdisplayskip{1pt}
    \alpha(\mathbf{W}_{1}^u, \mathbf{O}) = \alpha(\mathbf{W}_{1}^{u-1}, \mathbf{O}) + \log p(w_u|\mathbf{W}_{1}^{u-1}, \mathbf{O})
\end{equation}
where $\log p(w_u|\mathbf{W}_{1}^{u-1}, \mathbf{O})$ is the weighted sum of different log probabilities possibly delivered by the attention decoder, the acoustic encoder and the language models. 
In this work, log probability distribution provided by LF-MMI, namely $\log p_{\tt{MMI}}(w_u|\mathbf{W}_{1}^{u-1}, \mathbf{O})$, is additionally considered as a component of $\log p(w_u|\mathbf{W}_{1}^{u-1}, \mathbf{O})$. 
$\log p_{\tt{MMI}}(w_u|\mathbf{W}_{1}^{u-1}, \mathbf{O})$ can be derived from the first-order difference of $S^{\tt{pref}}_{\tt{MMI}}$: 
\begin{equation}
\setlength\abovedisplayskip{0.2cm}
\setlength\belowdisplayskip{0cm}
\label{eq_mmi_prefix_diff}
\log p_{\tt{MMI}}(w_u|\mathbf{W}_{1}^{u-1}, \mathbf{O}) = S^{\tt{pref}}_{\tt{MMI}}(\mathbf{W}_{1}^{u},\mathbf{O}) - S^{\tt{pref}}_{\tt{MMI}}(\mathbf{W}_{1}^{u-1},\mathbf{O})
\end{equation}
where $S^{\tt{pref}}_{\tt{MMI}}$ is defined as the summed probability of all hypotheses that start with $\mathbf{W}_{1}^{u}$. As shown in Eq.\ref{eq_mmi_prefix}, for any hypothesis $\mathbf{W}\in\mathbf{H}(\mathbf{W}_{1}^{u})$, we assume the partial hypothesis $\mathbf{W}_{1}^{u}$ is pronounced in first $t$ frames $\mathbf{O}_{1}^{t}$ while the complementary part of $\mathbf{W}_{1}^{u}$, namely $\mathbf{W}_{u+1}^{U}$, is pronounced in the remained $T-t$ frames $\mathbf{O}_{t+1}^{T}$. Additionally, as $\mathbf{O}$ and $\mathbf{W}$ are known, $\mathbf{W}_{u+1}^{U}$ is independent to $\mathbf{W}_{1}^{u}$ while $\mathbf{O}_{t+1}^{T}$ is independent to $\mathbf{O}_{1}^{t}$. Since each $t\in[1, T]$ could be valid and events with different $t$ are exclusive, probabilities are accumulated along t-axis, and the summed probability over the set $\mathbf{H}(\mathbf{W}_{1}^{u})$ is discarded (it is equal to 1). Finally, each element $P_{\tt{MMI}}(\mathbf{W}_{1}^{u}|\mathbf{O}_{1}^{t})$ is approximated by Eq.\ref{eq_mmi_graph}, where $\mathbb{G}_{num}(\mathbf{W}_{1}^{u})$ is the numerator graph built from $\mathbf{W}_{1}^{u}$.



\begin{equation}
\setlength\abovedisplayskip{-0.3cm}
\setlength\belowdisplayskip{0cm}
\label{eq_mmi_prefix}
\begin{aligned}
     S^{\tt{pref}}_{\tt{MMI}}&(\mathbf{W}_{1}^{u}, \mathbf{O}) = \log\sum_{\mathbf{W}\in \mathbf{H}(\mathbf{W}_{1}^{u})}P_{\tt{MMI}}(\mathbf{W}|\mathbf{O}) \\
     \approx& \log\sum_{t=1}^{T}\sum_{\mathbf{W}\in \mathbf{H}(\mathbf{W}_{1}^{u})} P_{\tt{MMI}}(\mathbf{W}_{1}^{u}|\mathbf{O}_{1}^{t})P_{\tt{MMI}}(\mathbf{W}_{u+1}^{U}|\mathbf{O}_{t+1}^{T}) \\
     =& \log\sum_{t=1}^{T}P_{\tt{MMI}}(\mathbf{W}_{1}^{u}|\mathbf{O}_{1}^{t}) \sum_{\mathbf{W}\in \mathbf{H}(\mathbf{W}_{1}^{u})}P_{\tt{MMI}}(\mathbf{W}_{u+1}^{U}|\mathbf{O}_{t+1}^{T}) \\
     =& \log\sum_{t=1}^{T}P_{\tt{MMI}}(\mathbf{W}_{1}^{u}|\mathbf{O}_{1}^{t}) \approx \log\sum_{t=1}^{T}\frac{P(\mathbf{O}_{1}^{t}|\mathbb{G}_{num}(\mathbf{W}_{1}^{u}))} {P(\mathbf{O}_{1}^{t}|\mathbb{G}_{den})}
\end{aligned}
\end{equation}



In Eq.\ref{eq_mmi_prefix}, the accumulation of probability along the t-axis seems computationally expensive. However, several properties of it could be considered to greatly alleviate this problem. First, unlike in the training stage, only the forward part of the \textit{forward-backward} algorithm is needed to calculate all terms in Eq.\ref{eq_mmi_prefix}. 
Second, the computation on the denominator graph is independent to the partial hypothesis $\mathbf{W}_{1}^{u}$, which could be done before the searching process and reused for any partial hypothesis proposed during beam search.
\vspace{-10pt}
\subsection{Beam Search for NTs}
For NTs, $S^{\tt{ali}}_{\tt{MMI}}$ is proposed to cooperate with the decoding algorithm ALSD\cite{alsd}. Note tuple ${(\textbf{W}_{1}^{u}, \delta_{t}(\textbf{W}_{1}^{u}), g_u)}$ as a hypothesis where $\textbf{W}_{1}^{u}$ is the output sequence (including no blank) with length $u$, $\delta_{t}(\textbf{W}_{1}^{u})$ is the hypothesis score and $g_u$ is the decoding state of prediction network. The subscript $t$ in $\delta_{t}(\textbf{W}_{1}^{u})$ means the hypothesis is aligned to first $t$ frames $\textbf{O}_{1}^{t}$. 

As hypotheses in NT decoding suggest explicit alignments, they can be evaluated by keeping $S^{\tt{ali}}_{\tt{MMI}}(\textbf{W}_{1}^{u},\textbf{O}_{1}^{t})=\log P_{\tt{MMI}}(\textbf{W}_{1}^{u}|\textbf{O}_{1}^{t})$ as a component of $\delta_{t}(\textbf{W}_{1}^{u})$ with a predefined weight $\beta$. Thus, once a new hypothesis is proposed (a new token or \textit{blank} is added), its score is computed recursively using Eq.\ref{mas_blank} and Eq.\ref{mas_token}. Similar to $S^{\tt{pref}}_{\tt{MMI}}$, we implement $S^{\tt{ali}}_{\tt{MMI}}$ by Eq.\ref{eq_mmi_graph} and emphasize the possibility to reuse the denominator scores during decoding. Note $S_{\tt{NT}}^{blk}$ and $S_{\tt{NT}}^{w_{u+1}}$ are the posteriors of \textit{blank} and token $w_{u+1}$ output by NT respectively.

\begin{equation}
\setlength\abovedisplayskip{-0.3cm}
\setlength\belowdisplayskip{0cm}
\label{mas_blank}
\begin{aligned}
     \delta_{t+1}(\textbf{W}_{1}^{u}) =& \delta_{t}(\textbf{W}_{1}^{u}) + S_{\tt{NT}}^{blk}(t,u) \\ 
     & +  \beta * (S^{\tt{ali}}_{\tt{MMI}}(\textbf{W}_{1}^{u},\textbf{O}_{1}^{t+1}) - S^{\tt{ali}}_{\tt{MMI}}(\textbf{W}_{1}^{u},\textbf{O}_{1}^{t}))
\end{aligned}
\end{equation}

\begin{equation}
\setlength\abovedisplayskip{0cm}
\setlength\belowdisplayskip{0cm}
\label{mas_token}
\begin{aligned}
     \delta_{t}(\textbf{W}_{1}^{u+1}) =& \delta_{t}(\textbf{W}_{1}^{u}) + S_{\tt{NT}}^{w_{u+1}}(t,u) \\ 
     & +  \beta * (S^{\tt{ali}}_{\tt{MMI}}(\textbf{W}_{1}^{u+1},\textbf{O}_{1}^{t}) - S^{\tt{ali}}_{\tt{MMI}}(\textbf{W}_{1}^{u},\textbf{O}_{1}^{t}))
\end{aligned}
\end{equation}

In each step when all proposed hypotheses are evaluated, scores of hypotheses that have identical $\textbf{W}_1^u$ but different alignment paths should be merged. But $S^{\tt{ali}}_{\tt{MMI}}$ should not participate in this process, since $S^{\tt{ali}}_{\tt{MMI}}$ directly assesses the validness of the aligned sequence pair $(\textbf{W}_1^u, \textbf{O}_1^t)$ and is the summed posterior of all alignment paths.   

\vspace{-10pt}
\subsection{Look-ahead Decoding Strategy}
A presumption of $S^{\tt{pref}}_{\tt{MMI}}$ and $S^{\tt{ali}}_{\tt{MMI}}$ is that the numerator graph could be composed for any partial hypothesis $\textbf{W}_1^u$. This is correct for languages like Mandarin since every proposed character from the neural decoder is also in the lexicon. However, it is incorrect for spelling languages like English, as a prefix of an English word is not always in the lexicon. E.g., \textit{speec}, as a prefix of word \textit{speech}, is not in the lexicon and the numerator graph cannot be compiled for it easily. 

Inspired by \cite{lookahead_lm}, we tackle this problem by computing a look-ahead score. For any partial hypothesis, we split it into two parts: word context $c$, which is the sequence of complete words in the front of the partial hypothesis, and prefix $p$, which is a prefix of a word at the end of the hypothesis. We denote each partial hypothesis as $\mathbf{W}_1^{u}=c\oplus p$. Thus, any log posterior $\log P_{\tt{MMI}}(\textbf{W}_{1}^{u}|\textbf{O}_1^t)$ of this partial hypothesis is formulated as below:
\begin{equation}
\setlength\abovedisplayskip{0cm}
\setlength\belowdisplayskip{0cm}
\label{eq_mmi_lookahead_prefix_score}
    \log P_{\tt{MMI}}(\textbf{W}_1^u|\textbf{O}_1^t) = \log\sum_{w\in\left\{p*\right\}} P_{\tt{MMI}}(c\oplus w|\mathbf{O}_1^t)
\end{equation}
where $\left\{p*\right\}$ indicates the set of all words in the lexicon that start with $p$. A special case is that the partial hypothesis consists of all complete words: $S^{\tt{pref}}_{\tt{MMI}}$ and $S^{\tt{ali}}_{\tt{MMI}}$ are computed like $p$ is the last complete word and $|\left\{p*\right\}|=1$. 

\begin{figure}[h]
\includegraphics[width=0.4\textwidth]{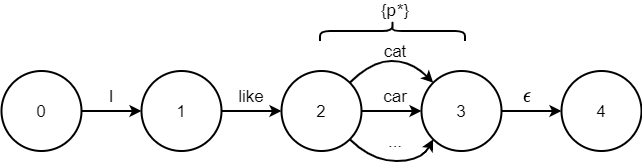}
\centering
\caption{Word FSA of partial hypothesis \textit{'I like ca'}. Word context $c=$\textit{'I like'} (arc 0$\rightarrow$1, 1$\rightarrow$2) Prefix $p=$\textit{'ca'} (arcs 2$\rightarrow$3). set $\left\{p*\right\}$ contains all words start with $p$ in the lexicon} 
\label{word_fsa}
\end{figure}

It seems that the summation in Eq. \ref{eq_mmi_lookahead_prefix_score} leads to heavy computation. However, all possible words $w\in\{p*\}$ could be converted into parallel arcs in a word FSA before compiling the numerator graph. E.g., a partial hypothesis \textit{'I like ca'} could be converted into a word FSA like in Fig \ref{word_fsa}, where the word context is arranged linearly while elements in $\left\{p*\right\}$ are converted into parallel arcs in the tail. This FSA is then composed with phone language model and HMM topology\cite{lfmmi_18} to derive the numerator graph for the forward computation. 

\subsection{MMI Rescoring}
We further propose a unified rescoring method called MMI Rescoring for both AEDs and NTs that are optimized with the LF-MMI criterion. Compared with using $S^{\tt{pref}}_{\tt{MMI}}$ and $S^{\tt{ali}}_{\tt{MMI}}$ in beam search, rescoring method is more computationally efficient. 

\begin{figure}
    \centering
    \includegraphics[width=7.5cm]{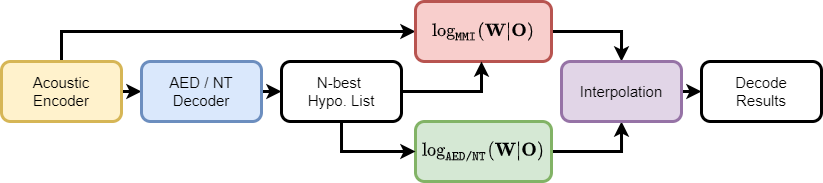}
    \caption{Diagram for MMI Rescoring. Interpolation from original posteriors and MMI posteriors are used for the final decision.}
    \label{fig_rescore}
\end{figure}

Assume the AED or NT system has been optimized by LF-MMI criterion before decoding. As illustrated in Fig.\ref{fig_rescore}, the N-best hypothesis list is firstly generated by beam search without LF-MMI criterion. Along this process, the log posterior of each hypothesis $\textbf{W}$, namely $\log P_{\tt{AED/NT}}(\mathbf{W}|\mathbf{O})$, is also calculated. Next, another log posterior for each hypothesis in the N-best hypothesis list, $\log P_{\tt{MMI}}(\mathbf{W}|\mathbf{O})$, is computed according to the LF-MMI criterion. Finally, the interpolation of the two log posteriors are calculated as follows:
\begin{equation}
\setlength\abovedisplayskip{0cm}
\setlength\belowdisplayskip{0cm}
\label{eq_rescore}
\begin{aligned}
    \log P(\mathbf{W}|\mathbf{O}) =& \lambda \cdot \log P_{\tt{AED/NT}}(\mathbf{W}|\mathbf{O}) \\
    &+ (1-\lambda ) \cdot \log P_{\tt{MMI}}(\mathbf{W}|\mathbf{O})
\end{aligned}
\end{equation}
where $\lambda$ is the weight of MMI Rescoring.
As LF-MMI criterion is applied to the acoustic encoder, MMI Rescoring could better emphasize the validness of hypotheses from the perspective of acoustics.

\noindent Moreover, since the denominator score $P(\mathbf{O}|\mathbb{G}_{den})$ is independent to the hypotheses, it could be considered as a constant for different hypotheses of a given utterance. Thus, only the numerator score needs to be calculated during MMI Rescoring:
\begin{equation}
\setlength\abovedisplayskip{0cm}
\setlength\belowdisplayskip{-0.5cm}
    \log P_{\tt{MMI}}(\mathbf{W}|\mathbf{O}) = \log P(\mathbf{O}|\mathbb{G}_{num}) - const
\end{equation}

\vspace{2pt}
\section{Experimental Results}
\label{chp_experiment}
\vspace{-8pt}
\subsection{Experimental Setup}
\textbf{Datasets}. We evaluate our method on Aishell-1 (178 hours, Mandarin), Aishell-2 (1000 hours, Mandarin) and Librispeech (960 hours, English) datasets. The modeling units for Mandarin and English are Chinese characters and BPE subwords respectively.

\noindent\textbf{Models and Optimization}. We adopt similar model architectures for experiments on all datasets. For AEDs, a Conformer encoder and a Transformer decoder (46M parameters) are used; while NTs consist of a Conformer encoder, an LSTM prediction network and an MLP joint network (89M parameters). All models are optimized by Noam\cite{noam} optimizer using 8 GPUs. 
We also adopt SpecAugment\cite{specaug} during training and average 10 checkpoints before evaluation. All experiments are implemented by Espnet\cite{espnet} and mainly follow the official settings\footnote{https://github.com/espnet/espnet/blob/master/egs/aishell/asr1/conf}. 

\noindent\textbf{Criteria}. We emphasize that our LF-MMI criterion adopts phone-level information so it is not fully End-to-End. 
All lexicons are from standard Kaldi recipes. The order of the phone language model used in the compilation of numerator and denominator graphs is 2. 
The HMM topology in our LF-MMI is the same as the CTC HMM topology in \cite{lfmmi_18}. 
Besides, we implement phone-level CTC with the same HMM topology and lexicon for comparison. 
Both LF-MMI and phone-level CTC are implemented with k2\footnote{https://github.com/k2-fsa/k2}. We also implement MBR training for NTs\cite{mbr_rnnt} with its original settings. 

\noindent\textbf{Decoding}. The beam size in all experiments is 10. Weights of MMI Prefix Score, MMI Alignment Score and MMI Rescoring are 0.3, 0.2, and 0.2 respectively.

\vspace{-8pt}
\noindent\subsection{Experimental Results}
We firstly present our results on Aishell-1 to provide a deep insight into our method. The effectiveness of the proposed method is further verified on two larger corpus (Aishell-2 and Librispeech).
\vspace{-8pt}
\subsubsection{Results of Aishell-1}

\begin{table}
\setlength{\abovecaptionskip}{0.cm}
\setlength{\belowcaptionskip}{-0.5cm}
\renewcommand\arraystretch{0.8}
\scriptsize
\begin{center}
\begin{tabular}{|c|l|c|c|c|}
\hline
\multirow{2}{*}{No.} &\multirow{2}{*}{System}  & Phone & \multicolumn{2}{|c|}{Aishell-1} \\
&   & Info. & dev & test  \\
\hline\hline 
\multicolumn{5}{|l|}{Literature} \\
\hline
  & Atten. \cite{lasresult}                           & \xmark   & 7.5       & 9.3       \\
  & Atten. + Char. CTC \cite{espnet}                  & \xmark   & 4.7       & 5.2       \\
  & Transd. \cite{espnet}                             & \xmark   & 4.3       & 4.8       \\
  & Non-Autoregressive Transformer\cite{nat}          & \xmark   & 5.6       & 6.3       \\
  & Chain (snowfall)                                  & \cmark   & -         & 6.3       \\ 
\hline\hline
\multicolumn{5}{|l|}{Attention-Based Encoder-Decoders (AEDs)} \\
\hline
1 & Atten. + Char. CTC\cite{lasctc}                  & \xmark   & 4.7       & 5.2        \\
2 & Atten. + LF-MMI Training\footnotemark[4]         & \cmark   & -         & -        \\
3 &     \ \ + MMI Prefix Score Decoding              & \cmark   & 4.6       & 5.2      \\
\hline
4 & Atten. + Char. CTC + Ph. CTC Training            & \cmark   & 4.6       & 5.1      \\
5 & Atten. + Char. CTC + LF-MMI Training             & \cmark   & \bf{4.5}  & 5.0      \\
6 &   \ \ + MMI Prefix Score Decoding                & \cmark   & \bf{4.5}  & 5.0      \\
7 &   \ \ + MMI Rescoring                            & \cmark    & \bf{4.5}  & \bf{4.9} \\
\hline\hline
\multicolumn{5}{|l|}{Neural Transducers (NTs)}                            \\
\hline  
8  & Transd.                                         & \xmark   & 4.4       & 4.8          \\
9  & Transd. + Ph. CTC Training                      & \cmark   & 4.8       & 5.2           \\
10 & Transd. + MBR Training\cite{mbr_rnnt}           & \xmark   & 4.7       & 5.1              \\
11 & Transd. + LF-MMI Training                       & \cmark   & 4.4       & 4.9            \\
12 & \ \ + MMI Alignment Score Decoding              & \cmark   & 4.3       & 4.7              \\
13 & \ \ + MMI Rescoring                             & \cmark   & 4.3       & 4.8             \\
\hline 
14 & Transd. + Char. CTC                             & \xmark   & 4.9       & 5.0               \\
15 & Transd. + Char. CTC + Ph. CTC Training          & \cmark   & 4.6       & 5.0             \\
16 & Transd. + Char. CTC + MBR Training\cite{mbr_rnnt}        & \xmark   & 4.7       & 5.2                \\
17 & Transd. + Char. CTC + LF-MMI Training           & \cmark   & 4.3       & 4.6               \\
18 &  \ \ + MMI Alignment Score Decoding             & \cmark   & 4.2  & 4.5            \\
19 &  \ \ \ \ + 4-gram Language Model Decoding       & \cmark   & \bf{4.1}  & \bf{4.4}   \\  
20 &  \ \ + MMI Rescoring                            & \cmark   & 4.2  & 4.5       \\
21 &  \ \ \ \ + 4-gram Language Model Decoding       & \cmark   & \bf{4.1}  & 4.5   \\
\hline
\end{tabular}
\end{center}
\caption{Experimental results on Aishell-1 dataset (CER\%)}
\label{table_aione}
\end{table}
\footnotetext[4]{Like \cite{lasctc}, decoding with only attention decoder cannot determine the ends of sentences accurately and results in unacceptable deletion errors.}
Table \ref{table_aione} shows the experimental results of the proposed method on Aishell-1 corpus. Several trends can be observed.
First, we adopt standard attention + character-level CTC and neural transducer as the baselines of AEDs (exp.1) and NTs (exp.8). 
Second, we claim that taking LF-MMI as an auxiliary criterion in training is beneficial if character-level CTC is used for regularization. Training with LF-MMI but without character-level CTC does not lead to a noticeable benefit (exp.2,3,11). However, with character-level CTC regularization, our training strategy pushes the baselines from 5.2\% to 5.0\% for AED (exp.5) and from 4.8\% for 4.6\% for NT (exp.17). 
Third, given the models trained with LF-MMI criterion (exp.5, 17), decoding with LF-MMI evidence in either beam search or rescoring can further improve the performance (exp.7, 18, 20), which emphasizes the necessity of the consistency between training and decoding.
Fourth, with a 4-gram character-level language model trained from the transcriptions, our model achieves the CER of 4.1\% and 4.4\% (exp.19, 21). To the best of knowledge, this is the state-of-the-art result of NT systems on Aishell-1.
Fifth, with identical HMM topology and lexicon, models trained with phone-level CTC (exp.4, 9, 15) are consistently worse than their LF-MMI counterparts (exp.5, 11, 17) or even show degradation compared with baselines (exp.9), which verifies that the effectiveness of our training strategy should be attributed to discriminative training rather than extra phone-level information. 
Finally, we also compare our method (exp.11, 17) with the character-level MBR criterion in NTs (exp.10, 16) but find that the MBR criterion does not achieve improvement. One possible explanation is that: the majority of the hypotheses proposed by the trained transducers are correct (the training corpus is well-fitted), which means the Bayesian Risk is equal to 0 and error signals provided by MBR are absent in most updates. In comparison, our method eschews the on-the-fly decoding process and provides error signals in every training step.

\vspace{-5pt}
\subsubsection{Results of Aishell-2 and Librispeech}
Due to the space limitation, we only report the NT results on Aishell-2 and AED results on Librispeech in table \ref{tab_aitwo} and table \ref{tab_libri} respectively. \\
\noindent\textbf{Aishell-2}. As shown in table \ref{tab_aitwo}, the trends of NT framework on Aishell-2 are similar to those of Aishell-1: (1) character-level CTC is still necessary for regularization (exp.2 vs. exp.5); (2) LF-MMI criterion is beneficial in training: up to 0.5\% absolute CER reduction is observed on \textit{test-ios} set (exp.1 vs. exp.5); (3) our decoding methods also achieve considerable improvement especially on \textit{test-mic} set (exp.5 vs. exp.6, 7).\\
\noindent\textbf{Librispeech}. As in table \ref{tab_libri}, our method is still beneficial during training. Adding LF-MMI as an auxiliary training criterion advances the WER of \textit{dev-clean} and \textit{test-other} datasets by 9.5\% and 5.6\% relatively while keeps other results unchanged (exp.1 vs. exp.4). In the decoding stage, however, degradation is observed in exp.3 and exp.6. We find that the MMI Prefix Score can hardly differentiate the repetitive tokens due to the time-axis probability accumulation in Eq.\ref{eq_mmi_prefix}, for which many deletion errors are observed in long and repetitive utterances. Since utterances in the two Mandarin datasets are comparatively shorter than those in Librispeech, this is rarely observed in those experiments. We leave this problem for future work. 

\begin{table}
\setlength{\abovecaptionskip}{0.cm}
\setlength{\belowcaptionskip}{-0.0cm}
\renewcommand\arraystretch{0.9}
\scriptsize
\begin{center}
\begin{tabular}{|c|l|c|c|c|}
\hline
\multirow{2}{*}{No.} & \multirow{2}{*}{System}& \multicolumn{3}{|c|}{Aishell-2-1000hrs}                          \\
 &  & ios & android & mic \\ 
\hline\hline 
1 & Transd.                                          & 5.9 & 6.7 & 6.5                                           \\
2 & Transd. + LF-MMI Training                        & 5.8 & 7.0 & 6.5                                           \\
3 & \ \ + MMI Alignment Score Decoding                       & 5.7 & 7.0 & 6.5                                                \\
4 & \ \ + MMI Rescoring                               & 5.7 & 6.9 & 6.5                                                 \\
\hline 
5 & Transd. + Char. CTC + LF-MMI Training                   & \textbf{5.4} & 6.6          & 6.5                             \\
6 &  \ \ + MMI Alignment Score Decoding                     & \textbf{5.4} & \textbf{6.5} & \textbf{6.3}                                     \\
7 &  \ \ + MMI Rescoring                           & \textbf{5.4} & 6.6          & 6.4                           \\
\hline
\end{tabular}
\end{center}
\caption{Neural Transducer results on Aishell-2 dataset (CER\%)}
\label{tab_aitwo}
\end{table}

\begin{table}
\setlength{\abovecaptionskip}{0.cm}
\setlength{\belowcaptionskip}{-0.5cm}
\renewcommand\arraystretch{0.9}
\scriptsize
\begin{center}
\begin{tabular}{|c|l|c|c|c|c|}
\hline
\multirow{2}{*}{No.} & \multirow{2}{*}{System}& \multicolumn{4}{|c|}{Librispeech-960hrs}  \\
 &  & d-c & d-o & t-c & t-o  \\ 
\hline\hline 
1 & Atten. + Char. CTC\cite{lasctc} & 2.1          & \textbf{5.0} & \textbf{2.2} & 5.3 \\
2 & Atten. + LF-MMI Training\footnotemark[4]        & -            & -            & -            & -   \\
3 & \ \ + MMI Prefix Score Decoding                & 2.2          & 5.4          & 2.6          & 5.4 \\
\hline
4 & Atten. + Char. CTC + LF-MMI Training & \textbf{1.9} & \textbf{5.0} & \textbf{2.2} & \textbf{5.0} \\
5 & \ \ + MMI Prefix Score Decoding         & 2.1          & 5.4          & 2.6          & 5.5 \\
6 & \ \ + MMI Rescoring               & \textbf{1.9} & \textbf{5.0} & \textbf{2.2} & 5.1 \\
\hline
\end{tabular}
\end{center}

\caption{Attention-Based Encoder-Decoders results on Librispeech dataset (WER\%)}
\label{tab_libri}
\end{table}

\vspace{-5pt}
\section{Conclusion}
This work is among the first works that integrate the LF-MMI criterion into End-to-End ASR frameworks. Unlike previous works, the proposed method consistently use LF-MMI criterion in both system training and decoding stages. In addition, the proposed method is compatible with both Attention-Based Encoder-Decoders and Neural Transducers. Experimental results suggest that our method achieves superior performance on three widely used ASR datasets.


\begin{scriptsize}
This paper was partially supported by the Shenzhen Science \& Technology Fundamental Research Programs (No: JCYJ20180507182908274 \& JSGG20191129105421211) and GXWD20201231165807007-20200814115301001.
\end{scriptsize}
\vfill\pagebreak

\bibliographystyle{IEEE}
\bibliography{refs}

\begin{thebibliography}{10}

\bibitem{e2easr_survey}
Dong Wang, Xiaodong Wang, and Shaohe Lv,
\newblock ``An overview of end-to-end automatic speech recognition,''
\newblock {\em Symmetry}, vol. 11, no. 8, pp. 1018, 2019.

\bibitem{las}
William Chan, Navdeep Jaitly, Quoc Le, and Oriol Vinyals,
\newblock ``Listen, attend and spell: A neural network for large vocabulary
  conversational speech recognition,''
\newblock in {\em 2016 IEEE International Conference on Acoustics, Speech and
  Signal Processing (ICASSP)}, 2016, pp. 4960--4964.

\bibitem{lasctc}
Shinji Watanabe, Takaaki Hori, Suyoun Kim, John~R Hershey, and Tomoki Hayashi,
\newblock ``Hybrid ctc/attention architecture for end-to-end speech
  recognition,''
\newblock {\em IEEE Journal of Selected Topics in Signal Processing}, vol. 11,
  no. 8, pp. 1240--1253, 2017.

\bibitem{rnnt}
Alex Graves,
\newblock ``Sequence transduction with recurrent neural networks,''
\newblock {\em arXiv preprint arXiv:1211.3711}, 2012.

\bibitem{ctc}
Alex Graves, Santiago Fern{\'a}ndez, Faustino Gomez, and J{\"u}rgen
  Schmidhuber,
\newblock ``Connectionist temporal classification: labelling unsegmented
  sequence data with recurrent neural networks,''
\newblock in {\em Proceedings of the 23rd international conference on Machine
  learning}, 2006, pp. 369--376.

\bibitem{mpe}
Karel Vesel{\`y}, Arnab Ghoshal, Luk{\'a}s Burget, and Daniel Povey,
\newblock ``Sequence-discriminative training of deep neural networks.,''
\newblock in {\em Interspeech}, 2013, vol. 2013, pp. 2345--2349.

\bibitem{povey_phd}
Daniel Povey,
\newblock {\em Discriminative training for large vocabulary speech
  recognition},
\newblock Ph.D. thesis, University of Cambridge, 2005.

\bibitem{smbr}
Brian Kingsbury,
\newblock ``Lattice-based optimization of sequence classification criteria for
  neural-network acoustic modeling,''
\newblock in {\em 2009 IEEE International Conference on Acoustics, Speech and
  Signal Processing}, 2009, pp. 3761--3764.

\bibitem{lfmmi_16}
Daniel Povey, Vijayaditya Peddinti, Daniel Galvez, Pegah Ghahremani, Vimal
  Manohar, Xingyu Na, Yiming Wang, and Sanjeev Khudanpur,
\newblock ``Purely sequence-trained neural networks for asr based on
  lattice-free mmi,''
\newblock in {\em Interspeech 2016}, 2016, pp. 2751--2755.

\bibitem{lfmmi_18}
Hossein Hadian, Hossein Sameti, Daniel Povey, and Sanjeev Khudanpur,
\newblock ``End-to-end speech recognition using lattice-free mmi,''
\newblock in {\em Proc. Interspeech 2018}, 2018, pp. 12--16.

\bibitem{mwer_las}
Rohit Prabhavalkar, Tara~N. Sainath, Yonghui Wu, Patrick Nguyen, Zhifeng Chen,
  Chung-Cheng Chiu, and Anjuli Kannan,
\newblock ``Minimum word error rate training for attention-based
  sequence-to-sequence models,''
\newblock in {\em 2018 IEEE International Conference on Acoustics, Speech and
  Signal Processing (ICASSP)}, 2018, pp. 4839--4843.

\bibitem{mbr_las}
Chao Weng, Jia Cui, Guangsen Wang, Jun Wang, Chengzhu Yu, Dan Su, and Dong Yu,
\newblock ``Improving attention based sequence-to-sequence models for
  end-to-end english conversational speech recognition.,''
\newblock in {\em Interspeech}, 2018, pp. 761--765.

\bibitem{mbr_rnnt}
Chao Weng, Chengzhu Yu, Jia Cui, Chunlei Zhang, and Dong Yu,
\newblock ``Minimum bayes risk training of rnn-transducer for end-to-end speech
  recognition,''
\newblock in {\em Proc. Interspeech 2020}, 2019.

\bibitem{mwer_rnnt}
Jinxi Guo, Gautam Tiwari, Jasha Droppo, Maarten Van~Segbroeck, Che-Wei Huang,
  and Stolcke,
\newblock ``Efficient minimum word error rate training of rnn-transducer for
  end-to-end speech recognition,''
\newblock in {\em Proc. Interspeech 2020}, 2020.

\bibitem{sa_mmi}
Naoyuki Kanda, Yashesh Gaur, Xiaofei Wang, Zhong Meng, Zhuo Chen, Tianyan Zhou,
  and Takuya Yoshioka,
\newblock ``Joint speaker counting, speech recognition, and speaker
  identification for overlapped speech of any number of speakers,''
\newblock in {\em Proc. Interspeech 2020}, 2020.

\bibitem{sa_mbr}
Naoyuki Kanda, Zhong Meng, Liang Lu, Yashesh Gaur, Xiaofei Wang, Zhuo Chen, and
  Takuya Yoshioka,
\newblock ``Minimum bayes risk training for end-to-end speaker-attributed
  asr,''
\newblock in {\em ICASSP 2021 - 2021 IEEE International Conference on
  Acoustics, Speech and Signal Processing (ICASSP)}, 2021, pp. 6503--6507.

\bibitem{alsd}
George Saon, Zoltán Tüske, and Kartik Audhkhasi,
\newblock ``Alignment-length synchronous decoding for rnn transducer,''
\newblock in {\em ICASSP 2020 - 2020 IEEE International Conference on
  Acoustics, Speech and Signal Processing (ICASSP)}, 2020, pp. 7804--7808.

\bibitem{lookahead_lm}
Takaaki Hori, Jaejin Cho, and Shinji Watanabe,
\newblock ``End-to-end speech recognition with word-based rnn language
  models,''
\newblock in {\em 2018 IEEE Spoken Language Technology Workshop (SLT)}. IEEE,
  2018, pp. 389--396.

\bibitem{noam}
Ashish Vaswani, Noam Shazeer, Niki Parmar, Jakob Uszkoreit, Llion Jones,
  Aidan~N Gomez, {\L}ukasz Kaiser, and Illia Polosukhin,
\newblock ``Attention is all you need,''
\newblock in {\em Advances in neural information processing systems}, 2017, pp.
  5998--6008.

\bibitem{specaug}
Daniel~S. Park, William Chan, Yu~Zhang, Chung-Cheng Chiu, Barret Zoph, Ekin~D.
  Cubuk, and Quoc~V. Le,
\newblock ``Specaugment: A simple data augmentation method for automatic speech
  recognition,''
\newblock {\em Interspeech 2019}, Sep 2019.

\bibitem{espnet}
Shinji Watanabe, Takaaki Hori, Shigeki Karita, Tomoki Hayashi, Jiro Nishitoba,
  Yuya Unno, Nelson {Enrique Yalta Soplin}, Jahn Heymann, Matthew Wiesner,
  Nanxin Chen, Adithya Renduchintala, and Tsubasa Ochiai,
\newblock ``{ESPnet}: End-to-end speech processing toolkit,''
\newblock in {\em Proceedings of Interspeech}, 2018, pp. 2207--2211.

\bibitem{lasresult}
Baiji Liu, Songjun Cao, Sining Sun, Weibin Zhang, and Long Ma,
\newblock ``Multi-head monotonic chunkwise attention for online speech
  recognition,''
\newblock {\em arXiv preprint arXiv:2005.00205}, 2020.

\bibitem{nat}
Xingchen Song, Zhiyong Wu, Yiheng Huang, Chao Weng, Dan Su, and Helen Meng,
\newblock ``Non-autoregressive transformer asr with ctc-enhanced decoder
  input,''
\newblock in {\em ICASSP 2021-2021 IEEE International Conference on Acoustics,
  Speech and Signal Processing (ICASSP)}. IEEE, 2021, pp. 5894--5898.

\end{thebibliography}
  \end{document}